\documentclass[runningheads]{llncs}

\PassOptionsToPackage{table}{xcolor}
\usepackage{eccv}

\usepackage{graphicx}
\usepackage{booktabs}
\usepackage{pifont}
\usepackage{bm}

\usepackage[accsupp]{axessibility}

\usepackage[hidelinks]{hyperref}
\usepackage{marvosym}

\newcommand{\refSec}[1]{Section {\ref{#1}}}

\makeatletter
\renewcommand\paragraph{\@startsection{paragraph}{4}{\z@}%
  {1.25ex \@plus 1ex \@minus .2ex}%
  {-1em}%
  {\normalfont\normalsize\bfseries}}
\makeatother

\newcommand{\ours}{MSVS-VAE}
\newcommand{\conv}{AVS-Conv}
\newcommand{\cmark}{\ding{51}}
\newcommand{\xmark}{\ding{55}}

\begin{document}

\title{MSVS-VAE: Multi-Scale Anchored VecSet for High-Fidelity 3D Reconstruction}

\titlerunning{MSVS-VAE}

\author{Dehao Hao\inst{1} \and
Kaiyi Zhang\inst{1} \and
Tanghui Jia\inst{2} \and
Xiangjun Gao\inst{1}\textsuperscript{\Letter} \and
Dongyu Yan\inst{3} \and
Weikai Chen\inst{4} \and
Zeyu Hu\inst{4} \and
Lingting Zhu\inst{4} \and
Yingda Yin\inst{4} \and
Runze Zhang\inst{4} \and
Li Yuan\inst{2} \and
Xin Wang\inst{4} \and
Long Quan\inst{1}}

\authorrunning{D.~Hao et al.}

\institute{
The Hong Kong University of Science and Technology, Hong Kong SAR, China\\
\email{xgaobq@connect.ust.hk}
\and
Peking University, Beijing, China
\and
The Hong Kong University of Science and Technology (Guangzhou), Guangzhou, China
\and
Lightspeed Studios, Shenzhen, China
}

\maketitle

\begingroup
\renewcommand{\thefootnote}{\Letter}
\footnotetext{Corresponding author}
\endgroup

\begin{abstract}
High-fidelity 3D generative modeling increasingly relies on the latent diffusion paradigm, where the reconstruction quality of the underlying 3D VAE becomes a primary bottleneck. Existing approaches largely follow two paradigms: \emph{sparse voxel-based} representations achieve strong reconstruction quality but incur significant memory and computational overhead, while \emph{set-based representations} are compact and continuous yet typically lag in fidelity due to latent sparsity and excessive global smoothness.
We propose \texttt{MSVS-VAE}, a hierarchical set-based VAE that closes this fidelity gap without sacrificing compactness.
Our key idea is to progressively densify \emph{anchored VecSet} latents via \emph{hierarchical point-shuffle upsampling}, increasing spatial capacity for fine-grained geometry modeling.
To efficiently decode from the densified hierarchy, we replace global cross-attention with \textit{\conv{}}, a geometry-aware local aggregation operator operating within local neighborhoods rather than the exhaustive latent set.
We further introduce \emph{multi-scale query decoding} to fuse coarse-to-fine latent features, where coarse scales provide stable global context, and fine scales refine localized geometry, reducing artifacts from overly local receptive fields.
Extensive experiments on Objaverse, ABO, and in-the-wild benchmarks demonstrate that \ours{} consistently outperforms prior set-based and voxel-based VAEs, delivering $\sim$10$\times$ faster decoding than prior set-based methods and $\sim$10$\times$ higher compactness than voxel-based baselines.

\end{abstract}

\begin{figure}[tb]
  \centering
  \includegraphics[width=1.0 \linewidth]{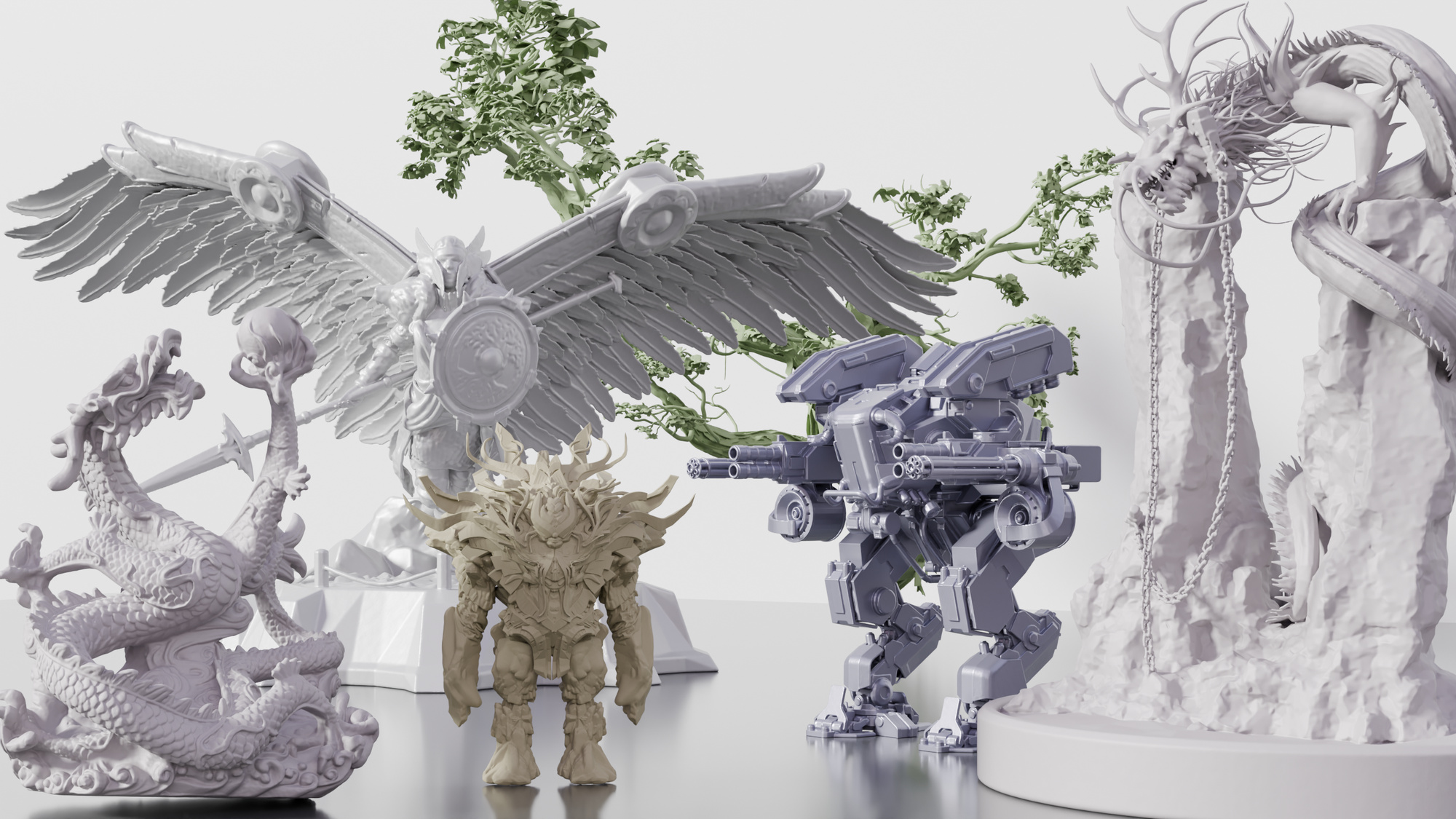}
  \caption{
  MSVS-VAE can achieve high-fidelity reconstruction using a compact vecset-based representation from a set of surface points.
  }
  \label{fig:teaser}
\end{figure}

\section{Introduction}
\label{sec:intro}

The pursuit of a high-fidelity yet scalable 3D representation remains a fundamental challenge of 3D generative modeling\cite{zhang2024clay, xiang2025structured, li2025triposg}. As the field gravitates toward the latent diffusion paradigm~\cite{rombach2022high, esser2021taming}, the performance of the 3D Variational Autoencoder (VAE)~\cite{kingma2013auto, zhang20233dshape2vecset} has become the primary bottleneck. A desirable 3D VAE must satisfy two conflicting requirements: it must be compact enough to allow efficient diffusion generative modeling, yet expressive enough to recover intricate geometric details during decoding.

Existing approaches largely fall into two representation paradigms.
\emph{Sparse voxel-based methods}\cite{li2025sparc3d, he2025sparseflex, ren2024xcube, xiang2025native, wu2025direct3d, xiang2025structured, zheng2023locally, hui2022neural} build upon dense grids with localized 3D convolutions or sparse attention.
Such representations naturally provide high spatial resolution and are effective at recovering sharp geometric details~\cite{he2025sparseflex, shen2023flexible}.
However, sparse voxel structures typically require a large number of active voxels, leading to high memory and computational cost~\cite{he2025sparseflex, chen2025ultra3d}.
Moreover, because the geometry is tied to a discrete grid, small prediction errors in the sparse structure may lead to brittle topology and a non-water-tight mesh~\cite{li2025sparc3d}.

In contrast, set-based representations \cite{lai2025lattice, hunyuan3d2025hunyuan3d, zhang20233dshape2vecset, li2025craftsman3d, li2025triposg, zhang2024clay, li2025step1x} encode shapes as compact latent sets and decode them through continuous query operators. This formulation is elegant and highly compact, while naturally enabling arbitrary-resolution surface queries. Recent advances, such as Lattice \cite{lai2025lattice}, significantly improve the reconstruction capability of set-based pipelines by introducing an \emph{anchored VecSet} representation—referred to as VoxSet—which augments VecSet tokens with explicit anchor positions (i.e., active voxel centers) to impose stronger spatial locality. Nevertheless, the reconstruction fidelity of set-based VAEs still lags behind voxel-based methods.

In this work, we argue that this fidelity gap arises from two fundamental limitations of existing VecSet-style representations.
First, decoding typically relies on global cross-attention between query points and the entire latent set, which implicitly imposes excessive \emph{global smoothness} and weakens the model’s ability to capture \emph{localized geometric details}. Second, VecSet representations are highly sparse—often containing tens of times fewer tokens than sparse voxel grids—which severely restricts the spatial resolution available during decoding. Although increasing the number of latent tokens could partially mitigate this issue, it incurs prohibitive computational overhead during both training and dense surface querying~\cite{lai2025unleashing}.

To address these limitations, we propose \textit{Multi-Scale VecSet VAE ({\ours{}})}, a hierarchical set-based VAE that pushes the reconstruction capability of VecSet representations to a new level.
Our key idea is to \emph{progressively densify the anchored VecSet} to enhance the capability for geometric details through a hierarchical latent upsampling mechanism, thereby increasing the spatial capacity available for modeling fine geometric details.
Specifically, we introduce a \textit{Hierarchical Point-Shuffle Upsampling} module~\cite{qian2021pu, shi2016real} that expands each anchored latent token into multiple localized sub-latents, progressively increasing the spatial density of the latent set while maintaining a compact overall representation. This densified anchored latent set equips VecSet-style VAEs with the spatial resolution necessary to recover fine-grained surface structures.

To efficiently operate on this expanded latent hierarchy, we further replace expensive global cross-attention with a \emph{geometry-aware local aggregation} operator, termed \textit{\conv{}}.
By querying only local neighborhoods in the latent space, \conv{} significantly reduces the computational cost of decoding while preserving the spatial inductive bias necessary for accurate surface reconstruction.
Finally, we introduce a \textit{Multi-Scale Query Decoding} module that aggregates information across multiple latent resolutions, allowing coarse latents to provide stable and long-range geometric context while fine latents capture localized geometric details\cite{muller2022instant, li2023neuralangelo, barron2021mip}. Fig.~\ref{fig:teaser} illustrates that our method reconstructs high-fidelity geometry from a compact anchored VecSet representation.

Together, these components enable set-based VAEs to achieve voxel-level \emph{reconstruction fidelity} while maintaining \emph{compactness} and improving the \emph{decoding efficiency}. Our main contributions are summarized as follows:

\begin{enumerate}

\item We propose \textbf{MSVS-VAE}, a novel set-based 3D VAE that enables high-fidelity 3D reconstruction while maintaining compact latent representations.
Our method achieves state-of-the-art reconstruction performance while providing $\sim$10$\times$ faster decoding than prior VecSet methods and $\sim$10$\times$ more compact representations than voxel-based baselines.

\vspace{5pt}
\item We introduce a \textit{Hierarchical Point-Shuffle Upsampling} module together with a \textit{Multi-Scale Query Decoding} module, which progressively increases the spatial resolution of the anchored VecSet and aggregates information across scales, significantly improving reconstruction quality while preserving global compactness.

\vspace{5pt}
\item We develop \textbf{\conv{}}, a geometry-aware local aggregation operator that replaces global cross-attention with efficient neighborhood-based decoding, substantially reducing computational cost while preserving fidelity.

\end{enumerate}

\section{Related Works}
Various 3D representations have been explored~\cite{wu2024direct3d, wang2023rodin, yang2019pointflow, wu2023fast, nichol2022point, jun2023shap}, but modern \emph{native} 3D generation largely centers on two paradigms: voxel-based and set-based representations.

\subsection{Voxel-based Representations}
Voxel-based methods scale 3D generation by introducing explicit sparse structures and multi-stage pipelines. XCube~\cite{ren2024xcube} generates hierarchical sparse voxels in a coarse-to-fine manner, enabling high effective resolutions.
Octfusion~\cite{xiong2025octfusion} compresses surface into an octree latents, and decodes  continuous SDFs with MPU~\cite{ohtake2005multi}. GaussianCube~\cite{zhang2024gaussiancube} constructs a grid-based 3DGS~\cite{kerbl20233d} representation via optimal transport.
Trellis~\cite{xiang2025structured} proposes structured latents that jointly model appearance and geometry, where geometry is represented via differentiable FlexiCubes~\cite{shen2023flexible}. SparseFlex~\cite{he2025sparseflex} further improves scalability with self-pruning and frustum-aware training, supporting high-resolution reconstruction and open-surface modeling. Direct3D-S2~\cite{wu2025direct3d} and SparC3D~\cite{li2025sparc3d} emphasize representation consistency by symmetrically encoding and decoding SDF-like volumetric signals (or sparse cubes). Concurrently, Trellis2~\cite{xiang2025native} and FaithC~\cite{luo2025faithful} adopts a dual-contouring-inspired~\cite{ju2002dual, chen2022neural} voxel representation to better capture complex topology and sharp features.
Despite strong reconstruction fidelity, voxel-based pipelines often co-define topology via explicit sparse structure and implicit field, making them sensitive to structure errors (e.g., holes).

\subsection{Set-based Representations}
Set-based methods~\cite{zhang20223dilg, zhang20233dshape2vecset} encode 3D surfaces into compact latent sets and decode SDFs via continuous, query-based operators. 3DILG~\cite{zhang20223dilg} queries latents through inverse-distance weighting, whereas VecSet~\cite{zhang20233dshape2vecset} replaces handcrafted interpolation with cross-attention, enabling single-stage generation. Michelangelo~\cite{zhao2023michelangelo} aligns the latent set with pretrained CLIP~\cite{radford2021learning} model for conditional generation, while Craftsman3D~\cite{li2025craftsman3d} refines the generated model with enhanced normals. Dora~\cite{chen2025dora} improves geometric quality via importance sampling to prioritize informative queries. OAT~\cite{deng2025efficient} introduces octree-based adaptive tokenization that encodes 3D shape into a variable-length set aligned with local geometric complexity. CLAY~\cite{zhang2024clay}, TripoSG~\cite{li2025triposg}, seed3d~\cite{seed2025seed3d}, and the Hunyuan3D series~\cite{lai2025hunyuan3d, hunyuan3d2025hunyuan3d, zhao2025hunyuan3d} further demonstrate the scalability of set-based latents, improving fidelity by scaling to larger models and datasets.
To push fidelity further, LATTICE~\cite{lai2025lattice} introduces a two-stage refinement scheme that re-injects explicit positional information into the VecSet diffusion framework, mirroring voxel-based pipelines to better recover fine-grained details. Nevertheless, set-based methods remain bottlenecked by VAE reconstruction quality, which still trails voxel-based alternatives~\cite{li2025sparc3d, wu2025direct3d, he2025sparseflex} in capturing fine-scale geometry.

\section{Preliminaries: Set-based 3D Representations}

Let a 3D shape be characterized by its continuous Signed Distance Function (SDF), denoted as $S(q) \in \mathbb{R}$ for any spatial query coordinate $q \in \mathbb{R}^3$. In contrast to methods that discretize space into explicit sparse grids, set-based representations encode the underlying 3D geometry into a discrete, permutation-invariant set of 1D latent vectors $\mathcal{Z} = \{z_i \in \mathbb{R}^C\}_{i=1}^M$. In this framework, $M$ specifies the number of latent tokens and $C$ defines the feature dimensionality. By abstracting the surface into a compact collection of latents, this representation naturally enables resolution-independent queries and provides a more flexible foundation for downstream generative modeling compared to grid-tied alternatives.

The VAE typically consists of an encoder that maps the input geometry into $\mathcal{Z}$, and a continuous, query-based decoder $\mathcal{D}$ that reconstructs the geometry. For any given spatial coordinate $q$, the SDF value is predicted by aggregating information from the global latent set:
\[
S(q) \approx \mathcal{D}(q, \mathcal{Z})
\]
In standard cross-attention-based frameworks (e.g., VecSet~\cite{zhang20233dshape2vecset}), this decoding is achieved by treating the spatial query $q$—often enriched via positional encoding~\cite{mildenhall2021nerf}—as the attention query, while the latent set $\mathcal{Z}$ serves as the keys and values. The extracted query feature is subsequently passed through a Multi-Layer Perceptron (MLP) to output the final SDF. During inference, it receives discrete grid queries and extracts mesh with iso-surface algorithms~\cite{lorensen1998marching, nielson2004dual}.

\section{Methodology}

In this section, we present \textbf{\textit{{\ours{}}}}, a hierarchical set-based VAE
for high-fidelity 3D reconstruction. We begin with a pipeline overview
in \refSec{m_1}, followed by detailed descriptions of the encoder in
\refSec{m_2} and the decoder in \refSec{m_3}.

\begin{figure}[htbp]
  \centering
  \includegraphics[width=1.0\linewidth]{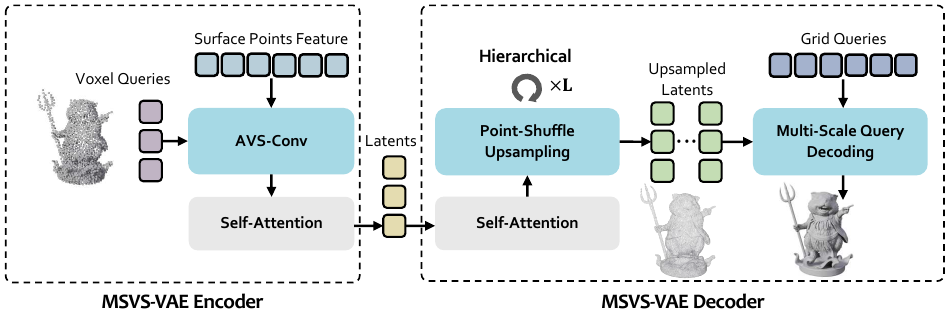}

  \caption{Overview of the proposed \textbf{\textit{{\ours{}}}}. The encoder extracts compact latents from surface points via \conv{} and self-attention blocks, while the decoder progressively upsamples the anchored latents and predicts SDF using a multi-scale query decoding module.}

  \label{fig:pipeline}
\end{figure}

\subsection{Overview: MSVS-VAE}
\label{m_1}

An overview of the proposed pipeline is illustrated in Fig.~\ref{fig:pipeline}. Given an input point cloud and a set of voxel queries, we first extract a compact latent set via {\textit{\conv{}}} from the input point cloud.  The resulting tokens are further refined by several self-attention blocks to produce anchored VecSet latents.
The compact anchored latent set is subsequently densified by {\textit{Hierarchical Point-Shuffle Upsampling}}, which progressively increases latent spatial density over multiple stages.
Finally, the {\textit{Multi-Scale Query Decoding}} module predicts SDF values at dense grid queries by aggregating features from neighboring latents across all upsampling scales.

\subsection{MSVS-VAE Encoder}
\label{m_2}

Given an input point cloud $\mathcal{X}=\{(x_n, a_n)\}_{n=1}^{N}$, where $x_n \in \mathbb{R}^3$ is the 3D position and $a_n \in \mathbb{R}^3$ is the surface normal, we first extract point-wise features ${f}_n = \Psi(a_n, x_n)$ using a lightweight PointNet.

Following Lattice~\cite{lai2025lattice}, we sample latent anchor points
$\mathcal{C} = \{c_m\}_{m=1}^M$ as voxel queries from the centers of active voxels intersected
by the mesh, obtained from an off-the-shelf generative model
(e.g., HY3D2.1\cite{hunyuan3d2025hunyuan3d}).

Treating $\mathcal{C}$ as the query set and $\mathcal{X}$ as the support
set, we apply \conv{} to aggregate point-wise features from the local
neighborhood of each query, producing latent tokens
$\mathcal{Z}=\{(c_m, z_m)\}_{m=1}^{M}$. These tokens are subsequently
refined by a stack of residual self-attention blocks. The \conv{} is defined as follows:

\paragraph{\conv{}. }
Different from previous VecSet-based methods~\cite{lai2025lattice, zhang20233dshape2vecset}, which rely on global cross-attention between latent queries and all input points, incurring $\mathcal{O}(MN)$ complexity and mixing information across the entire set. In contrast, we propose a geometry-aware local query-to-support operator whose cost scales more favorably with the number of queries while preserving locality.

\begin{figure}[tb]
  \centering
  \includegraphics[width=1.0 \linewidth]{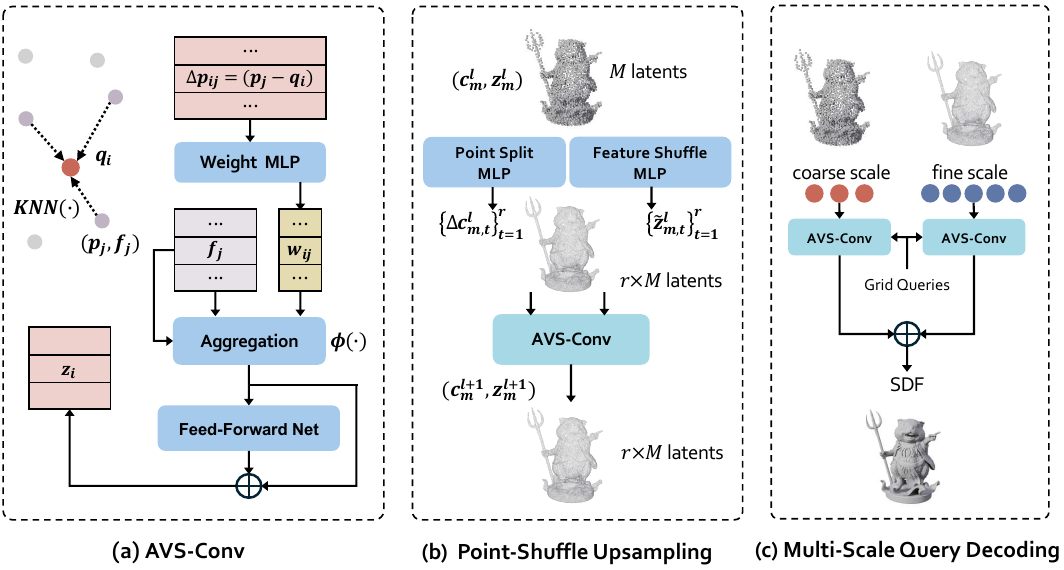}

  \caption{Illustrations of the three core modules in \ours{}.
(a)~\textbf{\textit{\conv{}}}: aggregates local support features for each
query via dynamic KNN-weighted convolution and a feed-forward residual block.
(b)~\textbf{\textit{Point-Shuffle Upsample}}: expands each latent into $r$ child
latents by splitting both its geometry and feature, followed by \conv{} refinement.
(c)~\textbf{\textit{Multi-Scale Query Decoding}}: fuses features independently
queried from coarse and fine latent scales to predict the SDF.}

  \label{fig:modules}
\end{figure}

As shown in Fig.~\ref{fig:modules}(a), for each query point $q_i$ in the query set, we retrieve its $K$ nearest neighbors $\mathcal{N}(i)$ from the support set under the Euclidean metric $\lVert p_j - q_i\rVert_2$. This explicit $K$NN retrieval defines a localized receptive field and eliminates the need for dense attention. Inspired by PointConv~\cite{wu2019pointconv}, we aggregate features over $\mathcal{N}(i)$ by predicting dynamic weights from relative offsets $\Delta p_{ij}=p_j-q_i$. Concretely, we compute the weight as:

\begin{equation}
\bm{w_{ij}} = \mathrm{MLP}(\gamma(\Delta p_{ij})), \qquad j\in\mathcal{N}(i),
\end{equation}

where $\gamma(\cdot)$ denotes a positional embedding applied to the relative offset $\Delta p_{ij}$. The resulting \conv{} is formulated as:

\begin{equation}
\bm{VC}(q_i,\mathcal{N}(i)) \;=\; \sum_{j\in\mathcal{N}(i)} w_{ij}\odot f_j,
\end{equation}

where $f_j$ is the support feature and $\odot$ denotes channel-wise multiplication. To improve optimization stability, we further incorporate a residual shortcut implemented as lightweight mean pooling over the same neighborhood:

\begin{equation}
\bm{\phi}(q_i;\mathcal{S}) = \bm{VC}(q_i,\mathcal{N}(i)) + {Mean}(q_i,\mathcal{N}(i)).
\end{equation}
Together, these components define a geometry-aware local aggregation operator $\phi(\cdot)$. Finally, we adopt a Transformer-style residual refinement with LayerNorm and a feed-forward network to produce the output feature $z_i$:
\begin{equation}
\bm{z_i} = \bm{\phi}(q_i;\mathcal{S}) + \mathrm{FFN}\left(\mathrm{LN}(\phi(q_i;\mathcal{S}))\right).
\end{equation}

Note that \conv{} is employed in both the encoder and decoder. In the decoder, it serves as a fundamental operator in both the Hierarchical Point-Shuffle Upsampling module and the Multi-Scale Query Decoding module. The main distinction lies in how the query and support sets are instantiated: in the encoder, the queries are VecSet anchors, and the support set comprises surface points; in the decoder, the queries are dense grid points, and the support set is the progressively upsampled anchored latent set.

\subsection{MSVS-VAE Decoder}
\label{m_3}

As discussed in \refSec{sec:intro}, the sparsity of VecSet representations—often tens of times more compact than sparse voxels—limits their ability to recover fine geometric details. Therefore, we propose a hierarchical VecSet latent upsampling module to enhance the capability of VecSet.

\paragraph{Hierarchical Point-Shuffle Upsampling.}
Given a compact latent set $\mathcal{Z}=\{(c_m, z_m)\}_{m=1}^{M}$ in Fig.~\ref{fig:modules}(b), we progressively increase the spatial density of VecSet latents while reducing per-latent channel capacity. Specifically, we apply $L$ upsampling stages to construct a hierarchy of latent supports
\[
\bm{\mathcal{Z}^{(l)}}=\{(c^{(l)}_m, z^{(l)}_m)\}_{m=1}^{M_l}, \quad l=0,\dots,L,
\]
where $\mathcal{Z}^{(0)}=\mathcal{Z}$ and $M_l$ grows exponentially with $l$.

For each stage  $l\!\rightarrow\!l{+}1$, we apply a point-shuffle upsampling step. Specifically, for each latent feature $z^{(l)}_m$ in $\mathcal{Z}^{(l)}$, we split it into $r$ sub-features $\{ \tilde z^{(l)}_{m,t}\}_{t=1}^{r}$ through a feature shuffle MLP with reduced channel dimension $C_{l+1}=C_l/2$. In parallel, we predict $r$ geometric offsets $\{\Delta c^{(l)}_{m,t}\}_{t=1}^{r}$ from $z^{(l)}_m$ using a point split MLP and form new centers
\[
\bm{c^{(l+1)}_{m,t}}=c^{(l)}_m+\Delta c^{(l)}_{m,t}, \qquad
\bm{z^{(l+1)}_{m,t}}=\tilde z^{(l)}_{m,t}.
\]
This transforms $M_l$ latents into $M_{l+1}=r \times M_l$ latents while keeping the total feature budget limited.

We then refine the upsampled latent set $\bm{\mathcal{Z}^{(l+1)}}$ using \conv{} over the updated centers ${c^{(l+1)}}$ and features ${z^{(l+1)}}$. Here, \conv{} is applied in a self-aggregation manner, where the query set and the support set are both the upsampled latent set $\bm{\mathcal{Z}^{(l+1)}}$. Moreover, the residual shortcut in $\phi(\cdot)$ is an identity connection on the latent features, rather than the mean pooling in the \ours{} encoder stage.

\paragraph{Multi-Scale Query Decoding.}

A fundamental distinction of our decoder from standard VecSet architectures is that we eliminate global cross-attention. In Fig.~\ref{fig:modules}(c), To predict the signed distance function (SDF) at an arbitrary spatial query $q_i\in\mathbb{R}^3$, we retrieve features from its $K$-NN latents using \text{\conv{}}. This local querying strategy avoids attending to all latents globally, and thus remains efficient even when the upsampled latent set becomes dense.

Relying only on the \emph{finest latent scale} can lead to high-frequency artifacts in smooth regions, since the receptive field becomes overly local. To mitigate this, we propose a multi-scale query fusion mechanism that aggregates information from both coarse and fine latent scales. Specifically, we query features from all $L$ progressive latent scales using separate \text{\conv{}} operators, concatenate the resulting features, and apply a lightweight fusion head:

\begin{equation}
\bm{h_i^{(l)}} = \mathrm{AVS}^{(l)}\!\left(q_i; \mathcal{Z}^{(l)}\right),
\qquad
\bm{h_i} = \mathrm{Fuse}\!\left([h_i^{(1)} \Vert \cdots \Vert h_i^{(L)}]\right)
\end{equation}
The final SDF value is predicted via
$\hat s_i = \mathrm{MLP}_{\mathrm{out}}(h_i)$.
This multi-scale design lets coarse latents supply \emph{stable, long-range geometric context}, while fine latents \emph{sculpt intricate local residuals}, enabling voxel-level fidelity within a continuous set-based representation.

\paragraph{Training Objectives.} Following standard practices~\cite{li2025triposg} for 3D VAEs, we supervise the reconstruction using the Truncated Signed Distance Function (TSDF) to prioritize near-surface geometry. The raw ground truth SDF $s$ is truncated and normalized as $\bar{s} = \text{clamp}(s / \tau, -1, 1)$, where the truncation distance is empirically set to $\tau = 1/128$. The overall training objective combines a geometric reconstruction loss and a Kullback-Leibler (KL) divergence regularizer:

\begin{equation}
\mathcal{L}_{\text{total}} = \lambda_{\text{recon}} \Big( \|\hat{s} - \bar{s}\|_1 + \|\hat{s} - \bar{s}\|_2^2 \Big) + \lambda_{\text{KL}} \mathcal{L}_{\text{KL}}
\end{equation}

where $\hat{s}$ denotes the predicted TSDF, and $\mathcal{L}_{\text{KL}}$ regularizes the latent tokens towards a standard normal distribution $\mathcal{N}(0, I)$.

\section{Experiments}

\subsection{Implementation Details}
We implement our \ours{} using PyTorch and Triton kernels. For nearest-neighbor search, we employ the BVH acceleration structure in cuBQL~\cite{cubql}, which significantly speeds up $K$-NN computation. We train our \ours{} on 400k high-quality meshes filtered from Objaverse~\cite{deitke2023objaverse}.

\paragraph{Training Strategy.}
To overcome the memory bottleneck of high-fidelity continuous surface learning, we adopt a progressive chunk-based training scheme. We first pre-train on full 3D assets with a fixed latent budget of $M{=}4096$ tokens to capture global topology and low-frequency geometry. We then fine-tune with localized random 3D crops: the same 4K tokens are allocated only to the cropped region (with a small spatial padding for context), effectively increasing local spatial resolution for high-frequency detail refinement. During the localized stage, we supervise SDF only on near-surface query points, as far-field SDF values depend on global topology beyond the sampled chunk and are therefore ill-posed under strictly local latents.

\paragraph{Training Settings.}
We train on 8 GPUs with a batch size of 4 per GPU for both stages. The global pre-training runs for 300k steps, followed by 100k steps of localized chunk fine-tuning. We use AdamW with a learning rate of $1\times10^{-4}$, $\beta=(0.9, 0.99)$, $\epsilon=10^{-6}$, and weight decay $10^{-2}$.

\subsection{Evaluation Settings}

\paragraph{Dataset.}
Following SparC3D~\cite{li2025sparc3d} and Dora~\cite{chen2025dora}, we curate an evaluation benchmark by selecting challenging shapes from Objaverse~\cite{deitke2023objaverse}, ABO~\cite{collins2022abo}, and in-the-wild Internet sources.
Our test set focuses on shapes with thin structures and complex topology, from which qualitative examples are sampled.
We further evaluate on the Level-3 and Level-4 subsets of Dora-Bench~\cite{chen2025dora}, which provide a widely used benchmark for challenging shapes with rich geometric details.

\paragraph{Baselines.}
We compare our method with the strongest prior \emph{set-based} approaches, including Lattice~\cite{lai2025lattice}, HY3D~2.1~\cite{hunyuan3d2025hunyuan3d}, and Dora~\cite{chen2025dora}, as well as state-of-the-art \emph{voxel-based} methods, including SparseFlex~\cite{he2025sparseflex}, SparC3D~\cite{li2025sparc3d}, and Direct3D-S2~\cite{wu2025direct3d}, where Lattice and SparC3D are our re-implementations following their original papers.

\paragraph{Metrics.}
We evaluate reconstruction quality using standard surface metrics. Mesh Distance (MD, $\times 10^4$) is computed as the bidirectional point-to-mesh distance.
We also report the F1-score ($\times 10^2$) at thresholds 0.01 and 0.001, which measures the harmonic mean of precision and recall based on whether bidirectional surface samples fall within the given distance threshold.
To better evaluate sharp geometric details, we additionally report sharp-edge-specific metrics on Dora-Bench, including one-sided Mesh Distance and F-score evaluated on ground-truth sharp points, denoted as S-MD and S-F, respectively.

We additionally benchmark \textbf{query latency} against prior cross-attention-based methods to validate the efficiency of our \conv{}.
Specifically, we report the average wall-clock time in milliseconds for \textbf{chunk-wise} point queries, using a fixed chunk size of 200k points per forward pass.
We evaluate three token budgets, 10k, 20k, and 40k, for both our method and Lattice.

\subsection{Evaluation Results}

\paragraph{Quantitative Comparisons.}

\begin{table}[tb]
\caption{VAE reconstruction evaluation on Objaverse, ABO, and in-the-wild benchmarks for voxel-based and set-based methods. \textit{$\dagger$ denotes our re-implementation.}}
  \label{tab:vae_recon}
  \centering
  \small
  \setlength{\tabcolsep}{3.5pt}
  \resizebox{\columnwidth}{!}{%
  \begin{tabular}{@{}l c ccc ccc ccc@{}}
    \toprule
    Method & \#tokens
      & \multicolumn{3}{c}{Objaverse}
      & \multicolumn{3}{c}{ABO}
      & \multicolumn{3}{c}{in-the-wild} \\
    \cmidrule(lr){3-5} \cmidrule(lr){6-8} \cmidrule(lr){9-11}
      &
      & MD $\downarrow$ & F1@0.01 $\uparrow$ & F1@0.001 $\uparrow$
      & MD $\downarrow$ & F1@0.01 $\uparrow$ & F1@0.001 $\uparrow$
      & MD $\downarrow$ & F1@0.01 $\uparrow$ & F1@0.001 $\uparrow$ \\
    \midrule

    \multicolumn{11}{@{}l}{\textit{Voxel-based methods}} \\
    \cmidrule(r){1-11}
    Direct3D-S2    & $\sim$74k & 3.301 & 99.955 & 95.100 & 2.091 & 99.987 & 97.181 & 2.771 & 99.928 & 97.020 \\
    SparC3D$^{\dagger}$          & $\sim$50k & 2.463 & 99.986 & \cellcolor{orange!20}{97.788} & 1.515 & 99.947 & 99.354 & \cellcolor{orange!20}{1.947} & 99.999 & 98.137 \\
    SparseFlex-512   & $\sim$50k & 3.745 & 99.867 & 93.182 & 2.017 & 99.958 & 97.449 & 4.404 & 99.915 & 90.634 \\
    SparseFlex-1024  & $\sim$210k & \cellcolor{orange!20}{2.165} & \cellcolor{orange!60}{99.999} & 97.327 & \cellcolor{orange!60}{1.092} & \cellcolor{orange!60}{100.00} & \cellcolor{orange!20}{99.902} & 2.014 & 99.998 & \cellcolor{orange!40}{98.944} \\

    \addlinespace[2pt]
    \cmidrule(r){1-11}
    \multicolumn{11}{@{}l}{\textit{Set-based methods}} \\
    \cmidrule(r){1-11}
    Dora             & 4k & 17.352 & 98.541 & 48.450 & 7.450 & 99.952 & 73.770 & 14.641 & 99.645 & 44.110 \\
    HY3D2.1         & 4k & 11.485 & 99.717 & 63.085 & 5.384 & 99.978 & 84.036 & 11.895 & 99.948 & 57.096 \\
    Lattice$^{\dagger}$          & 4k & 11.502 & 99.273 & 82.365 & 2.941 & 99.982 & 97.107 & 10.703 & 99.584 & 74.764 \\

    Lattice$^{\dagger}$          & 10k & 4.146 & 99.944 & 92.046 & 2.445 & 99.996 & 98.426 & 4.814 & 96.968 & 87.519 \\
    Lattice$^{\dagger}$          & 20k & 3.183 & 99.860 & 95.373 & 2.283 & 99.997 & 99.088 & 3.737 & 99.996 & 91.872 \\
    Lattice$^{\dagger}$          & 40k & 2.838 & 99.991 & 96.787 & 2.239 & 99.997 & 98.675 & 3.200 & \cellcolor{orange!60}{100.00} & 94.088 \\
    \ours{}& 4k & 5.644 & 99.714 & 86.779 & 1.885 & 99.956 & 98.902 & 4.589 & 99.984 & 87.207 \\

    \ours{}& 10k & 2.422 & 99.994 & 96.458 & 1.485 &  \cellcolor{orange!40}{99.999} & 99.843 & 2.416 & 99.999 & 95.842 \\
    \ours{}& 20k & \cellcolor{orange!40}{1.695} & \cellcolor{orange!40}{99.996} & \cellcolor{orange!40}{98.930} & \cellcolor{orange!20}{1.432} & \cellcolor{orange!40}{99.999} & \cellcolor{orange!60}{99.927} & \cellcolor{orange!40}{1.642} & \cellcolor{orange!60}{100.00} & \cellcolor{orange!20}{98.775} \\
    \ours{}& 40k & \cellcolor{orange!60}{1.395} & \cellcolor{orange!20}{99.992} & \cellcolor{orange!60}{99.621} & \cellcolor{orange!40}{1.216} & 99.995 & \cellcolor{orange!40}{99.924} & \cellcolor{orange!60}{1.231} & \cellcolor{orange!60}{100.00} & \cellcolor{orange!60}{99.726} \\
    \bottomrule
  \end{tabular}%
  }
\end{table}

\begin{table}[tb]
\caption{Evaluation on the Level-3 and Level-4 subsets of Dora-Bench for challenging shapes with sharp geometric details. MD and S-MD are scaled by $10^{-4}$; F and S-F are reported as percentages.}
  \label{tab:dora_bench_sharp_eval}
  \centering
  \scriptsize
  \setlength{\tabcolsep}{2.8pt}
  \renewcommand{\arraystretch}{0.92}

  \begin{tabular*}{\columnwidth}{@{\extracolsep{\fill}}l cccc cccc@{}}
    \toprule
    Method
      & \multicolumn{4}{c}{Level-3}
      & \multicolumn{4}{c}{Level-4} \\
    \cmidrule(lr){2-5} \cmidrule(lr){6-9}
      & MD$\downarrow$ & F$\uparrow$ & S-MD$\downarrow$ & S-F$\uparrow$
      & MD$\downarrow$ & F$\uparrow$ & S-MD$\downarrow$ & S-F$\uparrow$ \\
    \midrule

    Direct3D-S2
      & 3.99 & 95.6 & 5.37 & 91.3
      & 2.64 & 98.4 & 4.26 & 93.2 \\

    SparC3D
      & \cellcolor{orange!20}{1.65} & 99.0 & \cellcolor{orange!60}{2.80} & \cellcolor{orange!40}{96.4}
      & 2.14 & 97.4 & \cellcolor{orange!20}{3.22} & 94.4 \\

    SparseFlex
      & 1.74 & \cellcolor{orange!20}{99.1} & \cellcolor{orange!20}{3.21} & 95.6
      & \cellcolor{orange!20}{1.91} & \cellcolor{orange!20}{99.1} & \cellcolor{orange!60}{3.08} & \cellcolor{orange!40}{96.2} \\

    \addlinespace[2pt]
    \midrule
    \addlinespace[2pt]

    \ours{}-20K
      & \cellcolor{orange!40}{1.14} & \cellcolor{orange!40}{99.7} & 3.22 & \cellcolor{orange!40}{96.4}
      & \cellcolor{orange!40}{1.58} & \cellcolor{orange!40}{99.3} & 3.58 & \cellcolor{orange!20}{95.5} \\

    \ours{}-40K
      & \cellcolor{orange!60}{1.06} & \cellcolor{orange!60}{99.9} & \cellcolor{orange!40}{2.99} & \cellcolor{orange!60}{97.4}
      & \cellcolor{orange!60}{1.44} & \cellcolor{orange!60}{99.7} & \cellcolor{orange!40}{3.15} & \cellcolor{orange!60}{97.9} \\

    \bottomrule
  \end{tabular*}
  \vspace{-15pt}
\end{table}

As shown in Tab.~\ref{tab:vae_recon}, our method substantially and consistently surpasses the Lattice~\cite{lai2025lattice} baseline across all token budgets, benchmarks, and evaluation metrics, yielding a large improvement in geometry reconstruction quality.

Furthermore, using only \emph{20K} latent tokens, our method already outperforms competing voxel-based approaches on both Objaverse and in-the-wild benchmarks. It requires only \textbf{$\sim$10\%} (Ours: 20K \emph{vs} SparseFlex-1024: 210k) of the tokens used by SparseFlex-1024, and \textbf{$\sim$40\%} of those required by SparC3D and Direct3D-S2.
As shown in Tab.~\ref{tab:dora_bench_sharp_eval}, MSVS-VAE also achieves strong performance on Dora-Bench Level-3 and Level-4 subsets, especially on sharp-edge-specific metrics.
These quantitative comparisons demonstrate that our method achieves state-of-the-art geometric reconstruction quality with significantly more compact latents compared with the voxel-based method.

With only \emph{4K} tokens—the same budget as Dora and HY3D2.1—our method still outperforms them by a large margin on Objaverse (\emph{e.g. on Mesh Distance}, Ours: {5.644} \emph{vs} Dora: 17.352 \emph{vs} HY3D2.1: 11.485). The reported token numbers are averaged across all evaluation datasets.

\begin{figure}[!htbp]
  \centering
  \includegraphics[width=1.0 \linewidth]{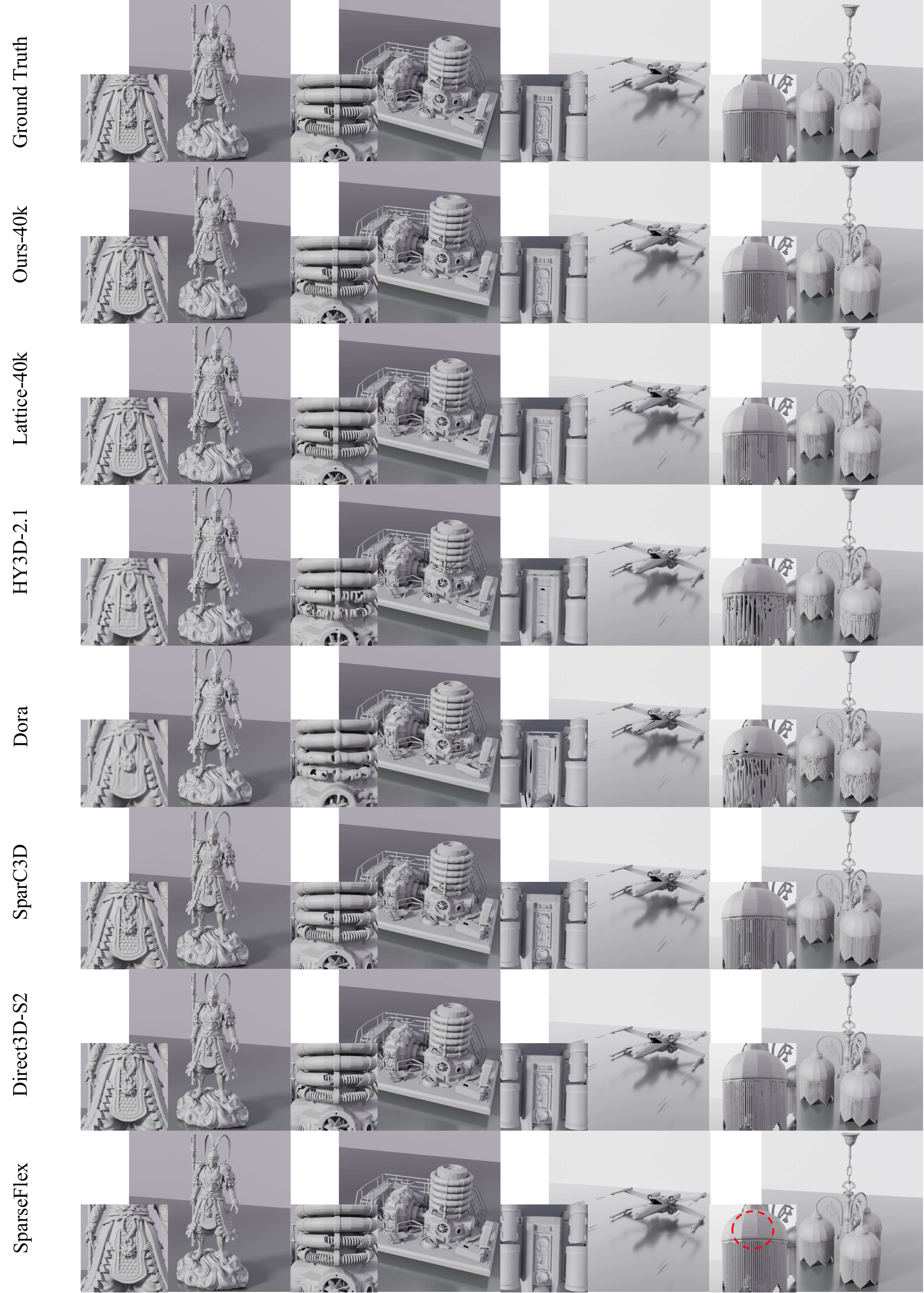}
  \caption{
  Qualitative comparison of VAE reconstruction. \textit{Best viewed with zoom-in}.
  }
  \label{fig:comparisons}
\end{figure}

\noindent \textbf{Qualitative Comparisons.}
Fig.~\ref{fig:comparisons} shows that our method faithfully reconstructs high-frequency surface details (\emph{see col.~1}), complex structures (\emph{see col.~2}), and delicate thin components (\emph{see col.~3--4}). Compared with \emph{set-based} methods, our approach achieves higher-fidelity reconstruction at both global and local levels, recovering intricate geometric details by leveraging the upsampled VecSet.

In addition, our method clearly outperforms most \emph{voxel-based}  (SparC3D and Direct3D-S2) baselines in reconstruction quality, producing sharper boundaries and substantially fewer surface sticking artifacts (\emph{see col.~3}).
Despite using only $\sim10\%$ of the tokens, our results remain competitive with or slightly better than SparseFlex-1024.
Moreover, our reconstructions are more topologically robust and watertight, avoiding the holes and broken parts observed in SparseFlex (\emph{see the red circle in col.~4; zoom in for details}). Overall, these results highlight a key advantage of set-based representations and demonstrate that our approach pushes the reconstruction capability of VecSet to the next level.

\emph{Discussion about SparseFlex:} Although the improvements over SparseFlex-1024 in both visual quality and quantitative metrics are modest, it is important to note that SparseFlex-1024 operates with a substantially denser latent space.
Specifically, it uses a spatial resolution of $256 \times 256 \times 256$ with a compression ratio of only $4\times4\times4$, whereas most voxel-based pipelines adopt a downsampling factor of around $8$ to keep the latent space manageable.
Such a dense latent space poses significant challenges for generative modeling due to the increased computational and modeling complexity.

\begin{table}[tb]
  \caption{Ablation study by component removal. Performance on Objaverse, ABO, and in-the-wild (MD$\downarrow$, F1@0.01$\uparrow$, F1@0.001$\uparrow$) after removing MSQ or PSU from the full model. Both modules contribute, and PSU is the most critical.}
  \label{tab:ablation_removal}
  \centering
  \small
  \setlength{\tabcolsep}{3.0pt}
  \resizebox{\columnwidth}{!}{%
  \begin{tabular}{@{}l cc ccc ccc ccc@{}}
    \toprule
    Setting & MSQ & PSU
      & \multicolumn{3}{c}{Objaverse}
      & \multicolumn{3}{c}{ABO}
      & \multicolumn{3}{c}{in-the-wild} \\
    \cmidrule(lr){4-6} \cmidrule(lr){7-9} \cmidrule(lr){10-12}
      & &
      & MD$\downarrow$ & F1@.01$\uparrow$ & F1@.001$\uparrow$
      & MD$\downarrow$ & F1@.01$\uparrow$ & F1@.001$\uparrow$
      & MD$\downarrow$ & F1@.01$\uparrow$ & F1@.001$\uparrow$ \\
    \midrule
    \rowcolor{gray!10}
    Full model            & \cmark & \cmark & \textbf{1.395} & \textbf{99.992} & \textbf{99.621} & \textbf{1.216} & \textbf{99.995} & \textbf{99.924} & \textbf{1.231} & \textbf{100.00} & \textbf{99.726} \\
    w/o MS-Query (MSQ)    & \xmark & \cmark & 1.690 & 99.959 & 99.322 & 1.439 & 99.967 & 99.890 & 1.626 & 99.977 & 99.416 \\
    w/o Upsample (PSU)     & \xmark & \xmark & 3.693 & 99.932 & 93.281 & 2.818 & 99.930 & 98.261 & 4.021 & 99.986 & 90.675 \\
    \bottomrule
  \end{tabular}%
  }
\end{table}

\begin{figure}[htbp]
  \centering
  \includegraphics[width=0.8\linewidth]{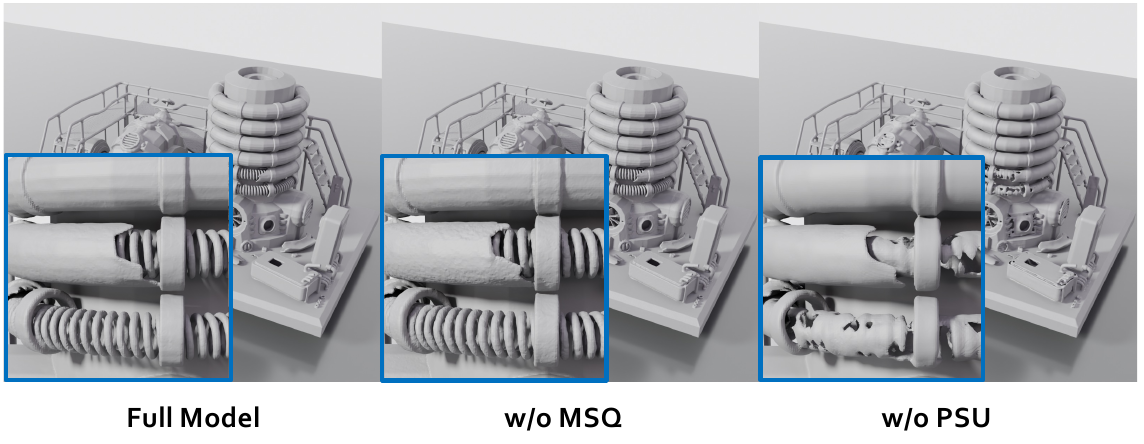}
    \vspace{-5pt}
  \caption{Qualitative ablation results of the proposed components. Removing MSQ results in noisier surfaces (\emph{Please zoom in for more details}), while removing PSU significantly degrades geometric reconstruction. The full model achieves smoother surfaces and more faithful reconstruction.}
  \label{fig:ablations1}
\end{figure}

\subsection{Ablation Studies}

\paragraph{Multi-scale Query Decoding (MSQ).}
As shown in Tab.~\ref{tab:ablation_removal}, multi-scale query decoding consistently improves over the single-scale variant, yielding lower MD and higher F-scores across all benchmarks. Fig.~\ref{fig:ablations1} further shows that MSQ produces smoother surfaces with fewer noise-induced artifacts, as fusing features across resolutions provides global context for structural consistency while preserving fine-scale cues for local details.

\paragraph{Hierarchical Point-Shuffle Upsampling (PSU).}
As shown in Tab.~\ref{tab:ablation_removal}, PSU is the most impactful component: removing it causes a drastic degradation in both MD and F-scores across all benchmarks. Fig.~\ref{fig:ablations1} further shows that PSU substantially improves geometric detail fidelity by progressively densifying the VecSet latents, thereby strengthening local geometric modeling and improving the recovery of high-frequency details. This module is therefore essential for pushing set-based representations toward voxel-level reconstruction quality.

\paragraph{Decoding Efficiency.}
Moreover, we compare query decoding efficiency against Lattice in Tab.~\ref{tab:query_latency}. Thanks to our geometry-aware local aggregation operator (\emph{\conv{}}), decoding latency increases much more slowly with the number of tokens, yielding consistent speedups over Lattice of $\sim$5.5$\times$/9.4$\times$/14.4$\times$ at 10k/20k/40k tokens, respectively. These results demonstrate that \emph{\conv{}} significantly reduces the computational cost of query decoding, enabling substantially faster decoding than existing \emph{set-based methods}.

\begin{table}[!t]
  \centering
  \scriptsize
  \setlength{\tabcolsep}{8pt}
  \renewcommand{\arraystretch}{1.1}
  \caption{\textbf{Query Latency Comparison.} We report the average latency (ms) for SDF querying with a chunk size of 200k. Our localized \textit{\conv{}} demonstrates significantly better scalability compared to the global-attention-based Lattice~\cite{lai2025lattice}.}
  \label{tab:query_latency}
  \begin{tabular}{@{}l cc c@{}}
    \toprule
    \# Latent Tokens & Lattice (ms) & \textbf{Ours (ms)} & \textbf{Speedup} \\
    \midrule
    10k & 792.21 & \textbf{143.75} & 5.5$\times$ \\
    20k & 1419.27 & \textbf{151.22} & 9.4$\times$ \\
    40k & 3068.05 & \textbf{213.29} & \textbf{14.4$\times$} \\
    \bottomrule
  \end{tabular}
  \vspace{-0.8\baselineskip}
\end{table}

\begin{figure}[htbp]
  \centering
  \includegraphics[width=1.0 \linewidth]{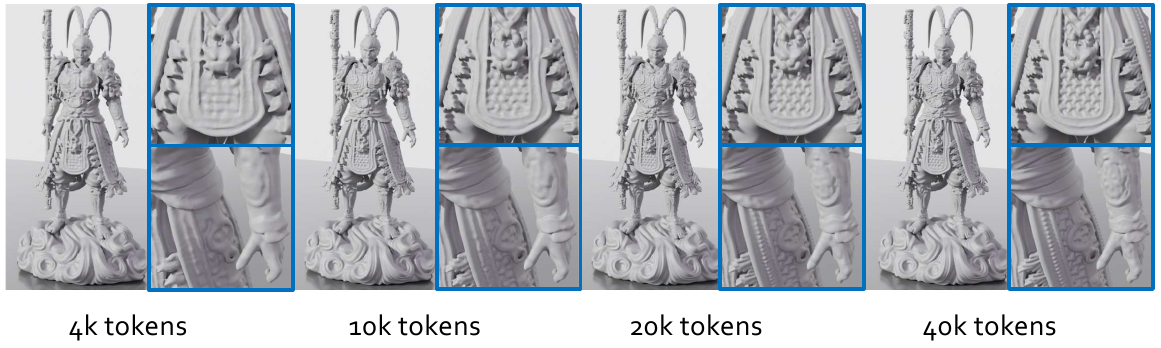}
  \vspace{-10pt}
  \caption{
  Qualitative test-time scaling with respect to token budget.
Increasing the number of tokens at inference time progressively improves reconstruction quality, enabling more faithful recovery of fine-grained geometric details.
  }
  \label{fig:tts}
\end{figure}

\begin{figure}[tb]
  \centering
  \includegraphics[width=0.6\linewidth]{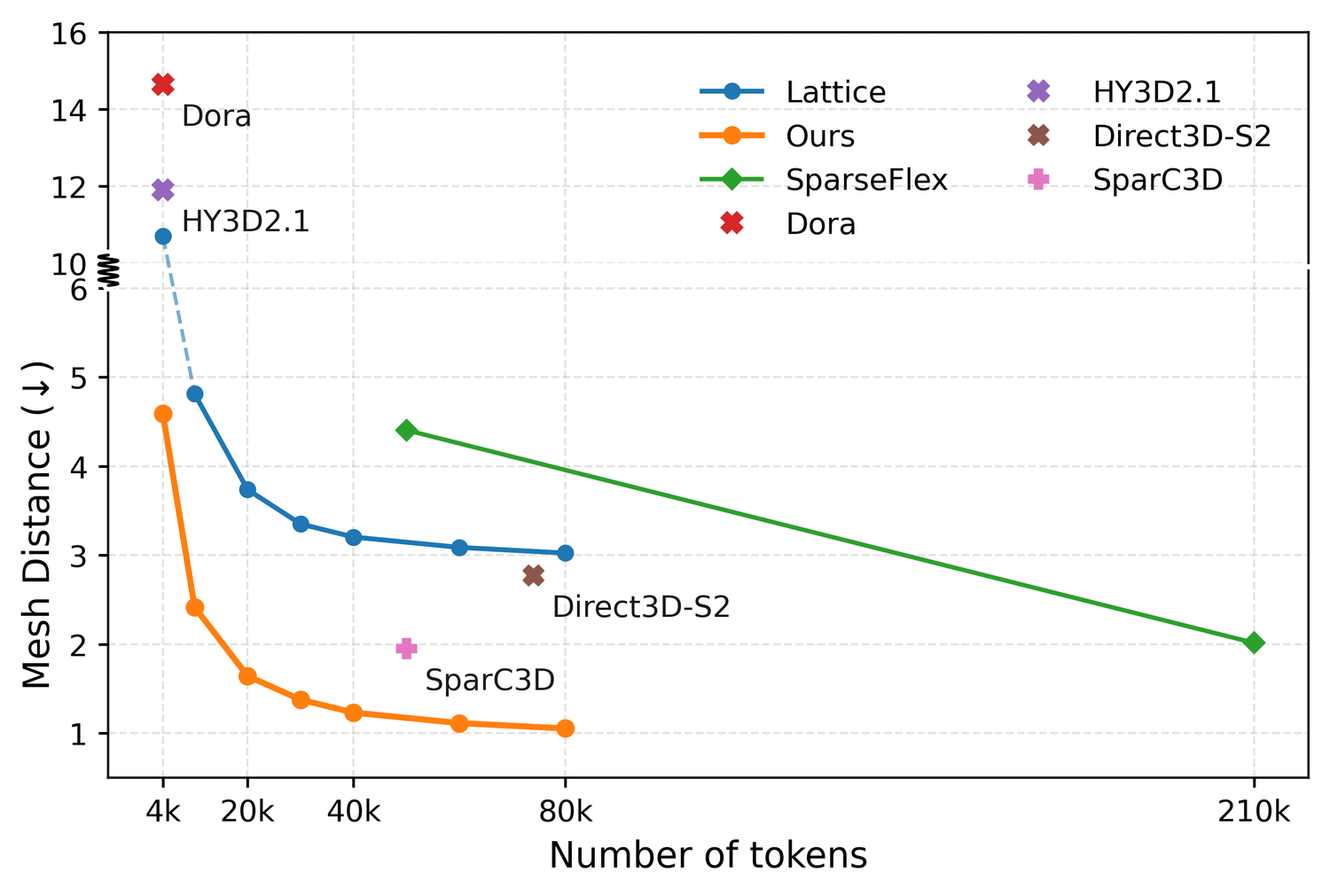}
  \vspace{-5pt}
  \caption{Reconstruction quality improves as the token budget increases. Our method consistently outperforms Lattice across token scales and achieves strong performance with far fewer tokens than voxel-based methods. (Evaluated on in-the-wild benchmark)}
  \label{fig:ablations2}
\end{figure}

\paragraph{Test-Time Scaling (TTS).}
We also evaluate the test-time scaling ability of our MSVS-VAE by varying the latent set size at inference time. We report reconstruction metrics with 4K, 10K, 20K, and 40K tokens on Objaverse, ABO, and in-the-wild. As shown in Tab.~\ref{tab:vae_recon}, increasing the number of tokens consistently improves reconstruction quality.
We additionally evaluate 30K, 60K, and 80K tokens on the in-the-wild benchmark and plot the results. Fig.~\ref{fig:ablations2} shows that, under the same token budget, our method consistently outperforms all competing approaches. Notably, using only \emph{20K} tokens, our model can achieve state-of-the-art performance, despite voxel-based methods operating with substantially larger latent spaces (e.g., 80K or 210K tokens).
Fig.~\ref{fig:tts} further illustrates test-time scaling: increasing the token budget consistently improves reconstruction quality.

\section{Conclusion}
We present \ours{}, a hierarchical set-based VAE that significantly improves the reconstruction capability of VecSet representations. By progressively densifying anchored VecSet latents through hierarchical point-shuffle upsampling and decoding them with multi-scale query aggregation, our approach effectively increases the spatial capacity of compact latent sets while maintaining efficient decoding. In addition, the proposed \conv{} operator replaces global cross-attention with geometry-aware local aggregation, substantially reducing computational cost. Extensive experiments demonstrate that our method achieves state-of-the-art reconstruction performance across multiple benchmarks, while remaining significantly more compact and efficient than voxel-based alternatives. These results show that compact set-based representations can approach voxel-level reconstruction fidelity, offering a promising direction for scalable 3D generative modeling.

\paragraph{Limitations.} Although \ours{} achieves state-of-the-art performance on geometry reconstruction, several limitations remain.
(1) Extracting sharp features remains challenging under a set-based continuous implicit formulation.
(2) The decoding efficiency is still behind voxel-based methods.
(3) Extending \ours{} to texture modeling is an important direction for future work.

\bibliographystyle{splncs04}
\bibliography{main}

\end{document}